# Distributed Parameter Estimation via Pseudo-likelihood

Qiang Liu                                                                                       QLIU1@UCI.EDU
Alexander Ihler                                                                               IHLER@ICS.UCI.EDU
Department of Computer Science, University of California, Irvine, CA 92697, USA

## Abstract

Estimating statistical models within sensor networks requires distributed algorithms, in which both data and computation are distributed across the nodes of the network. We propose a general approach for distributed learning based on combining local estimators defined by pseudo-likelihood components, encompassing a number of combination methods, and provide both theoretical and experimental analysis. We show that simple linear combination or max-voting methods, when combined with second-order information, are statistically competitive with more advanced and costly joint optimization. Our algorithms have many attractive properties including low communication and computational cost and "any-time" behavior.

## 1. Introduction

Wireless sensor networks are becoming ubiquitous, with applications including ecological monitoring, health care, and smart homes. Traditional centralized approaches to machine learning are not well-suited to sensor networks, due to the sensors' restrictive resource constraints. Sensors have limited local computing, memory, and power, and their wireless communication is expensive in terms of power consumption. These constraints make centralized data collection and processing difficult. Fault-tolerance and robustness are also critical features.

Graphical models are a natural framework for distributed inference in sensor networks (e.g., Cetin et al., 2007). However, most learning algorithms are not distributed, requiring centralized data processing and storage. In this work, we provide a general framework for distributed parameter estimation, based on combining local and inexpensive estimators.

This paper is organized as follows. Section 2 sets up background on graphical models for sensor networks and learning algorithms. In Section 3, we propose a framework for distributed learning based on intelligently combining results from disjoint local estimators. We give theoretic analysis in Section 4 and experiments in Section 5. We discuss related work in Section 6 and finally conclude the paper in Section 7.

## 2. Background

### 2.1. Graphical models for sensor networks

Consider a graphical model of a random vector $x = [x_1, \ldots, x_p]$ in exponential family form,

$$p(x|\theta) = \exp(\theta^T u(x) - \log Z(\theta)), \qquad (1)$$

where $\theta = \{\theta_\alpha\}_{\alpha \in \mathcal{I}}$ and $u(x) = \{u_\alpha(x_\alpha)\}_{\alpha \in \mathcal{I}}$ are vectors of the same size, and $\theta^T u(x)$ is their inner product. $\mathcal{I}$ is a set of variable indexes and $u_\alpha(x_\alpha)$ are local sufficient statistics. $Z(\theta)$ is the partition function, which normalizes the distribution. The distribution is associated with a Markov graph $G = (V, E)$, with node $i \in V$ denoting a variable $x_i$ and edge $(ij) \in E$ representing that $x_i$ and $x_j$ co-appear in some $\alpha$, that is, $\{i, j\} \subset \alpha$. Let $\beta_i = \{\alpha \in \mathcal{I} | i \in \alpha\}$ be the set of $\alpha$ that includes $i$. In pairwise graphical models, $\mathcal{I} = E \cup V$.

To model a sensor network, we represent the $i$-th sensor's measurement by $x_i$, and assume that the communication links between sensors are identical to the Markov graph $G$, that is, sensor $i$ and $j$ have communication link if and only if $(ij) \in E$. Assume that $n$ independent samples $X = [x^1, \ldots, x^n]$ are drawn from a true distribution $p(x|\theta^*)$. Due to memory and communication constraints on sensors, the data are stored locally within the network: each sensor stores only data measured by itself and its neighbors, that is, $X_{\mathcal{A}(i)} = [x^1_{\mathcal{A}(i)}, \ldots, x^n_{\mathcal{A}(i)}]$, where $\mathcal{A}(i) = \{i\} \cup \mathcal{N}(i)$ and $\mathcal{N}(i)$ is the neighborhood of node $i$. The goal is to design a distributed algorithm for estimating the true $\theta^*$, with minimum communication and low, balanced local computational costs at the sensor nodes.

**Notation.** Unless specified otherwise, we take $\mathbb{E}(\cdot)$, $\text{var}(\cdot)$, and $\text{cov}(\cdot)$ to mean the expectation, variance,

Distributed Parameter Estimation via Pseudo-likelihoodand covariance matrix under the true distribution $p(x|\theta^*)$. For a likelihood function $\ell(\theta; x)$, $\nabla \ell(\theta)$ and $\nabla^2 \ell(\theta)$ denote the gradient and Hessian matrix w.r.t. $\theta$, where we suppress the dependence of $\ell(\theta, x)$ on $x$ for compactness. We use "hat" accents to denote empirical average estimates, e.g., $\hat{\ell}(\theta, X) = \frac{1}{n}\sum_{k=1}^{n} \ell(\theta, x^k)$.

### 2.2. M-estimators

M-estimators are a broad class of parameter estimators; an M-estimator with criterion function $\ell(\theta; x)$ is

$$\hat{\theta} = \arg\max_{\theta} \hat{\ell}(\theta; X).$$

In this paper we assume that $\ell(\theta, x)$ is continuous differentiable and has a unique maximum. If $\mathbb{E}[\nabla \ell(\theta^*)] = 0$, then under mild conditions standard asymptotic statistics (van der Vaart, 1998) show that $\hat{\theta}$ is asymptotically consistent and normal, that is, $\sqrt{n}(\hat{\theta} - \theta^*) \rightsquigarrow \mathcal{N}(0, V)$, with asymptotic variance (Godambe, 1960)

$$V = H^{-T} J H^{-1},$$

where $J = \text{var}(\nabla \ell(\theta^*))$ is the Fisher information matrix and $H = -\mathbb{E}(\nabla^2 \ell(\theta^*))$ is the expected Hessian matrix. $\ell$ is said to be *information-unbiased* (Lindsay, 1988) if $J = H$. In this case, we have $V = H^{-1} = J^{-1}$, i.e., the asymptotic variance equals the inverse Fisher information matrix or Hessian matrix. Let $s$ be a random vector with $s = H^{-1} \nabla \ell(\theta^*, x)$. An important intuition for asymptotic analysis is that $\hat{\theta} \approx \theta^* + \frac{1}{\sqrt{n}} s$ at the large sample limit, so that the asymptotic variance can be rewritten as $V = \text{var}(s)$.

Empirically, one can assess the quality of an M-estimator by estimating its asymptotic covariance; this can be done by approximating the $\mathbb{E}(\cdot)$ and $\text{var}(\cdot)$ with their empirical counterparts, and $\theta^*$ with $\hat{\theta}$, e.g., the asymptotic variance is estimated by $\hat{V} = \hat{H}^{-1} \hat{J} \hat{H}^{-1}$, where $\hat{J} = \frac{1}{n}\sum_{k=1}^{n}(\nabla \ell(\hat{\theta}; x^k))(\nabla \ell(\hat{\theta}; x^k))^T$ and $\hat{H} = -\frac{1}{n}\sum_{k=1}^{n} \nabla^2 \ell(\hat{\theta}; x^k)$. If $\ell$ is information-unbiased, only the Fisher information $J$ need be calculated, avoiding calculating the second derivatives. In practice, these variance estimators perform well only when the parameter dimension is much smaller than the sample size; they are usually not directly applicable to practically sized problems. In this work, we show that by splitting the global estimator into low-dimensional local estimators, we can use covariance estimation on the local estimators to provide important information for combining them.

### 2.3. MLE and MPLE

The maximum likelihood estimator (MLE) is the most well-known M-estimator; it maximizes the likelihood,

$$\ell_{\text{mle}}(\theta; x) = \log p(x|\theta).$$

The MLE is asymptotically consistent and normal, and achieves the Cramér-Rao lower bound (is asymptotically at least as efficient as any unbiased estimator). Unfortunately, the MLE is often difficult to compute, because the likelihood involves the partition function $Z(\theta)$, which is hard to evaluate for general graphical models (Wainwright & Jordan, 2008).

The maximum pseudo-likelihood estimator (MPLE) (Besag, 1975) provides a computationally efficient alternative to MLE. The pseudo-likelihood is defined as

$$\ell_{\text{mple}}(\theta; x) = \sum_{i=1}^{p} \log p(x_i | x_{\mathcal{N}(i)}; \theta_{\beta_i}), \qquad (2)$$

where due to the Markov property, each conditional likelihood component only depends on $\theta_{\beta_i}$, the parameters incident to $i$, and on $X_{\mathcal{A}(i)}$, the data available to sensor $i$. MPLE remains asymptotically consistent and normal, but is usually statistically less efficient than MLE – a sacrifice for computational efficiency. However, cases exist in which the MPLE is also statistically more favorable than MLE, e.g., when the model is misspecified (e.g., Liang & Jordan, 2008).

There is a weaker version of MPLE, well known in sparse learning (e.g., Ravikumar et al., 2010), that disjointly maximizes the single conditional likelihood (CL) components in MPLE, and then combines the overlapping components using some simple method such as averaging. Very recently, the disjoint MPLE has started to attract attention in distributed estimation (Wiesel & Hero, 2012), by observing that the conditional likelihoods define local estimators well suited to distributed computing.

**Our work.** We address the problem of distributed parameter learning within a paradigm motived by MPLE and disjoint MPLE, in which the sensor nodes locally calculate their own inexpensive local estimators, whose results are communicated to nearby sensors and combined. We provide a more general framework for combining the local estimators, including weighted linear combinations, a max-voting method, and more advanced joint optimization methods. Powered by asymptotic analysis, we propose efficient methods to set optimal weights for the linear and max combination methods, and provide a comprehensive comparison of the proposed algorithms. Surprisingly, we show that the simple linear and max combination methods, when leveraged by well-chosen weights, are able to outperform joint optimization in some cases. In particular, the max-voting method performs well on "degree-unbounded" graph structures, such as stars or scale free networks, that are difficult for many existing methods. In addition, we show that the joint

MPLE can be recast into a sequence of disjoint MPLE combinations via the alternating direction method of multipliers (ADMM), and we show that, once it is initialized properly, interrupting the iterative algorithm at any point provides "correct" estimates; this leads to an *any-time* algorithm that can flexibly trade off performance and resources, and is robust to interruptions such as sensor failure. Finally we provide extensive simulation to illustrate our theoretical results.

## 3. A Distributed Paradigm

For each sensor $i$, let $\ell^i_{\text{local}}(\theta_{\beta_i}; x_{\mathcal{A}(i)})$ be a criterion function that depends only on local data $X_{\mathcal{A}(i)}$ and the parameter sub-vector $\theta_{\beta_i}$. This defines a M-estimator that is efficient to compute locally by sensor $i$,

$$\hat{\theta}^i_{\beta_i} = \arg\max_{\theta_{\beta_i}} \hat{\ell}^i_{\text{local}}(\theta_{\beta_i}; X_{\mathcal{A}(i)}). \qquad (3)$$

We assume that $\mathbb{E}(\nabla \ell^i_{\text{local}}(\theta^*_\beta)) = 0$ and that (3) has a unique maximum, which guarantee that $\hat{\theta}^i_{\beta_i}$ is asymptotically consistent and normal under standard technical conditions. Further, assume $\cup_i \beta_i = \mathcal{I}$, so that each parameter component is covered by at least one local estimator and a valid global estimator can be constructed by combining them.

Although our results apply more generally, in this work we mainly take $\ell^i_{\text{local}}(\theta_{\beta_i}; x_{\mathcal{A}(i)}) = \log p(x_i | x_{\mathcal{N}(i)}; \theta_{\beta_i})$, which satisfies the conditions listed above. Moreover, such $\ell^i_{\text{local}}(\theta_{\beta_i})$ are information unbiased, i.e., $V^i_{\text{local}} = (J^i_{\text{local}})^{-1} = (H^i_{\text{local}})^{-1}$. One can estimate the asymptotic variance by $\hat{V}^i_{\text{local}} = (\hat{J}^i_{\text{local}})^{-1}$, where $\hat{J}^i_{\text{local}}$ involves calculating the covariance of the gradient statistics and is efficient once $|\beta_i|$ is relatively small.

If a parameter $\theta_\alpha$ is shared by multiple sensors, performance can be boosted by combining their information. We propose two types of consensus methods, generalizing disjoint MPLE and MPLE respectively.

### 3.1. One-Step Consensus.

For each parameter $\theta_\alpha$, let $\hat{\boldsymbol{\theta}}_\alpha = \{\hat{\theta}^i_\alpha | i \in \alpha\}$ be the collection of estimates given by the sensors incident to $\alpha$. The goal is to construct a combined estimator $\hat{\theta}_\alpha$ as a function of $\hat{\boldsymbol{\theta}}_\alpha$. Probably the simplest method is averaging, i.e., $\hat{\theta}_\alpha = \frac{1}{|\alpha|} \sum_{i \in \alpha} \hat{\theta}^i_\alpha$. Unfortunately, as we show in the sequel, this simple approach usually performs poorly, in part because it weights all the estimators equally and the worst estimator may greatly degrade the overall performance. Thus, it would be helpful to weight the estimators by their quality.

Let $\hat{w}^i_\alpha$, as a function of $X_{\mathcal{A}(i)}$ and $\hat{\theta}^i_{\text{local}}$, be an empirical measure of the quality of the $i$-th local estimator for estimating parameter $\theta_\alpha$ – for example, $\hat{w}^i$ could be a function of $\hat{V}^i_{\text{local}}$. We introduce two methods to combine the estimators based on weight $\hat{w}^i$:

**linear consensus:**

$$\hat{\theta}^{\text{linear}}_\alpha = \sum_{i \in \alpha} \hat{w}^i_\alpha \hat{\theta}^i_\alpha / \sum_{i \in \alpha} \hat{w}^i_\alpha, \qquad (4)$$

**max consensus:**

$$\hat{\theta}^{\max}_\alpha = \hat{\theta}^{i_0}_\alpha, \qquad \text{where } \hat{w}^{i_0}_\alpha \geq \hat{w}^i_\alpha \text{ for all } i \in \alpha, \qquad (5)$$

where the linear consensus takes a soft combination of the local estimators, while the max consensus votes on the best one. It should be noted that the max consensus can be treated as a special linear consensus whose weights are taken be to be zero, except on one local estimator. However, as we show later, the max consensus has some attractive properties making it an efficient algorithm for many problems.

We prove that linear and max consensus are asymptotically consistent and normal, and provide their asymptotic variance. We also discuss the optimal setting of the weights, in the sense of minimizing the asymptotic mean square error. Remarkably, we show that the optimum weights, particularly for the max consensus, are surprisingly easy to estimate, making one-step methods competitive to more advanced consensus methods.

### 3.2. Joint Optimization via ADMM

A more principled way to ensure consensus is to solve a joint optimization problem,

$$\max_{\theta^i_{\beta_i}, \bar{\theta}} \sum_{i=1}^n \hat{\ell}^i_{\text{local}}(\theta^i_{\beta_i} | X_{\mathcal{A}(i)}) \text{ s.t. } \theta^i_{\beta_i} = \bar{\theta}_{\beta_i} \quad \text{for all } i \quad (6)$$

where we maximize the sum of $\hat{\ell}^i_{\text{local}}$ under the constraint that all the local estimators should be consistent with a global $\bar{\theta}$; this exactly recovers the joint MPLE method in (2) when $\ell^i_{\text{local}}$ are the conditional likelihoods. In this section, we derive a distributed algorithm for (6) that can be treated as an iterative version of the linear consensus introduced above.

Our algorithm is based on the alternating direction method of multipliers (ADMM), which is well suited to distributed convex optimization (Boyd et al., 2011), particularly distributed consensus (Bertsekas & Tsitsiklis, 1989).

For notation, let $f^i(\theta^i_{\beta_i}) = -\hat{\ell}^i_{\text{local}}(\theta^i_{\beta_i} | X_{\mathcal{A}(i)})$. We introduce an augmented Lagrangian function for (6),

$$\sum_{i=1}^p \left\{ f^i(\theta^i_{\beta_i}) + \lambda^i_{\beta_i}{}^T (\theta^i_{\beta_i} - \bar{\theta}_{\beta_i}) + \sum_{\alpha \in \beta_i} \frac{\rho^i_\alpha}{2} |\theta^i_\alpha - \bar{\theta}_\alpha|^2 \right\},$$

where $\lambda^i_{\beta_i}$ are Lagrange multipliers of the same size as $\theta^i_{\beta_i}$ and $\rho^i_{\beta_i}$ are positive penalty constants. Performing an alternating direction procedure on the augmented Lagrangian yields the ADMM algorithm:

$$\theta^i_{\beta_i} \leftarrow \arg\min\{f^i(\theta^i_{\beta_i}) + \lambda^{i\,T}_{\beta_i}\theta^i_{\beta_i} + \sum_{\alpha\in\beta_i}\frac{\rho^i_\alpha}{2}||\theta^i_\alpha - \bar\theta_\alpha||^2\}$$

$$\bar\theta_\alpha \leftarrow \sum_{i\in\alpha}\rho^i_\alpha\theta^i_\alpha / \sum_{i\in\alpha}\rho^i_\alpha, \quad \forall\alpha\in\mathcal{I}$$

$$\lambda^i_\alpha \leftarrow \lambda^i_\alpha + \rho^i_\alpha(\theta^i_\alpha - \bar\theta_\alpha), \quad \forall\alpha\in\beta_i,$$

This update has an intuitive statistical interpretation. First, $\theta^i_{\beta_i}$ can be treated as a posterior MAP estimation of the parameter subject to a Gaussian prior with mean $(\bar\theta_{\beta_i} - \lambda^i_{\beta_i}/\rho^i_{\beta_i})$, which biases the estimate towards the average value; $\bar\theta$ is then re-evaluated by taking a linear consensus of the local estimators. Thus, the joint optimization can be recast into a sequence of linear consensus steps. Given this connection, it is reasonable to set $\rho^i_\alpha$ to be the weights of linear consensus, that is, $\rho^i_\alpha = \hat w^i_\alpha$ and initialize $\bar\theta$ to be the corresponding one-step estimator. Since linear consensus estimators are asymptotically consistent, we have

**Theorem 3.1.** *If we set $\bar\theta$ to be asymptotically consistent and $\lambda^i_{\beta_i} = 0$ in the initial step of ADMM, then $\bar\theta$ remains asymptotically consistent at every iteration.*

Therefore, one can interrupt the algorithm and fetch a "correct" answer at any iteration, giving a flexible *anytime* framework that can not only save on computation and communication, but is also robust to accidental failures, such as battery depletion.

## 4. Asymptotic Analysis

In this section, we give an asymptotic analysis of our methods, by which we provide methods to optimally set the weights of linear and max consensus. For notational convenience, we embed the local estimator $\hat\theta^i_{\beta_i} = \arg\max \hat\ell^i_{\text{local}}(\theta_{\beta_i}, X)$ into a (possibly degenerate) estimator of the whole parameter vector $\hat\theta^i = \arg\max \hat\ell^i(\theta, X)$, by setting $\hat\theta^i_\alpha = 0$ for $\alpha \notin \beta_i$. Denote by $V^i$ the asymptotic variance of the extended estimator, with $V^i_{\text{local}}$ on its $\beta_i \times \beta_i$ sub-matrix and zero otherwise. Similarly, let $H^i$ extend $H^i_{\text{local}}$, and $s^i$ extend $s^i_{\beta_i} \stackrel{def}{=} H^i_{\text{local}}{}^{-1}\nabla\ell^i_{\text{local}}(\theta^*_{\beta_i})$. Our results will reflect the intuition that $\hat\theta^i \approx \theta^* + \frac{1}{\sqrt{n}}s^i$ at the large sample limit.

For our results, we generalize to a matrix extension of the linear consensus (4), defined as

$$\hat\theta^{\text{matrix}} = (\sum_i \hat W^i)^{-1}\sum_i \hat W^i \hat\theta^i, \qquad (7)$$

where $\hat W^i$ are matrix weights that are non-zero only on the $\beta_i \times \beta_i$ sub-matrices; we require that $(\sum_i \hat W^i)^{-1}$ is invertible. Note that the matrix extension is not directly suitable for distributed implementation, since it involves a global matrix inverse, but it will provide performance bounds for linear and max consensus and has close connection to joint optimization estimators.

**Theorem 4.1** (Linear Consensus). *Assume $\hat W^i \stackrel{p}{\to} W^i$ and $\sum_i W^i$ is an invertible matrix. Then $\hat\theta^{\text{matrix}}$ in (7) is asymptotically consistent and normal, with an asymptotic variance of $\text{var}\big[(\sum_i W^i)^{-1}\sum_i W^i s^i\big]$.*

Assume $H = \sum_i H^i$ is invertible, then the joint optimization consensus $\hat\theta^{\text{joint}} = \arg\max\sum_i \hat\ell_i(\theta, x)$ is a non-degenerate estimator of the full parameter vector $\theta$. It turns out $\hat\theta^{\text{joint}}$ is asymptotically equivalent to a matrix linear consensus with weights $\hat W^i = \hat H^i$:

**Corollary 4.2.** *$\hat\theta^{\text{matrix}}$ in (7) with $\hat W^i = \hat H^i$ has asymptotic variance of $\text{var}[(\sum_i H^i)^{-1}\sum_i \nabla\ell^i(\theta^*)]$, which is the same as that of $\hat\theta^{\text{joint}}$.*

For max consensus estimators, we have

**Theorem 4.3.** *The $\hat\theta^{\max}$ in (5) is asymptotically consistent. Further, for any $\alpha \in \mathcal{I}$, if $\hat w^i_\alpha \stackrel{p}{\to} w^i_\alpha$ and $w^{i_0}_\alpha > \max_{i\in\alpha, i\ne i_0} w^i_\alpha$, then $\hat\theta^{\max}_\alpha$ is asymptotically normal, with asymptotic variance equal to $V^{i_0}_{\alpha,\alpha}$.*

### 4.1. Optimal Choice of Weights

In this section, we consider the problem of choosing the optimal weights, in the sense of minimizing the asymptotic mean square error (MSE). Note that $\mathbb{E}(||\hat\theta - \theta^*||^2) \to \frac{1}{n}\text{tr}V$ as $n \to +\infty$, where $\text{tr}(V)$ is the trace of the asymptotic covariance matrix, and so the problem can be reformed to minimize $\text{tr}(V)$. In the following, we discuss the optimal weights for the linear and max consensus separately.

**Weights for Max Consensus.** The greedy nature of max consensus makes optimal weights relatively easy:

**Proposition 4.4.** *For the max consensus estimator $\hat\theta^{\max}$ as defined in (5), the weight $w^i_\alpha = 1/V^i_{\alpha,\alpha}$ achieves minimum least square error asymptotically.*

In practice, we can estimate the optimal weights simply by $\hat w^i_\alpha = 1/\hat V^i_{\alpha,\alpha}$, which makes max consensus feasible in practice.

**Weights for Linear Consensus.** By Theorem 4.1, the optimal weights for matrix linear consensus solve

$$\min_{W^i}\text{tr}[\text{var}(\sum_i W^i s^i)] \quad \text{s.t.} \quad \sum_i W^i = \mathbf{1}, \qquad (8)$$

where $W^i$ are non-zero only on the $\beta_i \times \beta_i$ submatrix and $\mathbf{1}$ denotes the identity matrix of the same size as $W^i$. Solving (8) is difficult in general, but if $s^i$ are pairwise independent and $\hat{\theta}^i$ are information-unbiased, the weights $W^i = H^i$, asymptotically equivalent to $\hat{\theta}^{\text{joint}}$ as shown in Corollary 4.2, achieves optimality.

**Proposition 4.5.** *Assume $\hat{\theta}^i$ are information-unbiased. If $\text{cov}(s^i, s^j) = 0$ for all $i \neq j$, then $\hat{\theta}^{\text{joint}}$ achieves the optimum MSE as defined in (8).*

This implies that if the estimators are independent or weakly correlated, the joint optimization estimator $\hat{\theta}^{\text{joint}}$ is guaranteed to perform no worse than the linear and max consensus methods (both suboptimal to the best matrix consensus). However, in the case that the local estimators are strongly correlated (usually the case in practice), there is the chance that linear or max consensus, with properly chosen weights, can outperform the joint optimization method.

On the other hand, when $W^i$ in (8) are constrained to be diagonal matrices, reducing to a set of vector weights $w_\alpha^i$, the optimization becomes easier. Let $w_\alpha = \{w_\alpha^i\}_{i \in \alpha}$ and $V_\alpha$ be an $|\alpha| \times |\alpha|$ matrix with $V_\alpha^{ij} = \text{cov}(s_\alpha^i, s_\alpha^j)$, i.e., $V_\alpha$ is the covariance matrix between the local estimators on parameter $\theta_\alpha$. Then,

**Proposition 4.6.** *For linear consensus estimator $\hat{\theta}^{\text{linear}}$ as defined in (4), the weights $w_\alpha = V_\alpha^{-1} e$, where $e$ is a column vector of all ones, achieves the minimum asymptotic least square error.*

In other words, the optimal vector weights for linear consensus equal the column sums of $V_\alpha^{-1}$. In practice, these weights can be estimated by $\hat{w}_\alpha = \hat{V}_\alpha^{-1} e$, where $\hat{V}_\alpha^{ij} = \frac{1}{n} \sum_{k=1}^n s_\alpha^i(x^k) \cdot s_\alpha^j(x^k)$, and $s_\alpha^k(x^k) = (\hat{H}^i)^{-1} \nabla \ell^i(\hat{\theta}^i; x^k)$. In the sensor network setting, calculating $V_\alpha^{ij}$ requires a secondary communication step in which the sensors pass $\{s_\alpha^i(x^k)\}_{k=1}^n$ to their neighbors. Note that this communication step may be expensive if the number of data $n$ is large (although one can pass a subset of samples to get a rougher estimate).

It is interesting to compare to the optimal weights for max consensus in Proposition 4.4, where no communication step is required. This is because max consensus fundamentally ignores the correlation structure, while linear consensus must account for it. Some further useful insights arise by considering the cases of extremely weak or strong correlations.

**Proposition 4.7.** *If $\text{cov}(s^i, s^j) = 0$, $\forall i \neq j$, then the $\hat{\theta}^{\text{linear}}$ as defined in (4) achieves the lowest asymptotic MSE with weights $w_\alpha^i = 1/V_{\alpha,\alpha}^i$.*

This suggests that $\hat{w}_\alpha^i = 1/\hat{V}_{\alpha,\alpha}^i$, which is optimal for max consensus selection, might also be a reasonable choice for linear consensus. However, the independence assumption is always violated in practice. To see what happens when the estimators are strongly correlated, consider the opposite extreme, in which the local estimators are deterministically correlated:

**Proposition 4.8.** *If $s^i$ $(i = 1, \ldots, p)$ are deterministically positively correlated, i.e., there exists a random vector $s^0$, and constants $v_\alpha^i \geq 0$, such that $s_\alpha^i = v_\alpha^i s_\alpha^0$, then the optimal vector weights $\{w_\alpha^i\}$ for linear consensus, under the constraint $w_\alpha^i \geq 0$, is $w_\alpha^i = 1$ if $v_\alpha^i \leq v_\alpha^j$ for any $j \in \alpha$ and $w_\alpha^i = 0$ if otherwise.*

Since linear consensus with 0-1 weights reduces to max consensus, this result suggests that the optimal max consensus is not much worse than the optimal linear consensus when the estimators are strongly positively correlated. In practice, we find that the local estimators defined by conditional likelihoods are always positively correlated, justifying max consensus in practice.

### 4.2. Illustration on One Parameter Case

In this section, we illustrate our asymptotic results in a toy example, providing intuitive comparison of our algorithms. Assume $\theta$ is a scale parameter, estimated by two information-unbiased estimators $\hat{\theta}^i = \arg\max \ell^i(\theta)$ $(i = 1, 2)$. Let $h^i = -\mathbb{E}(\nabla^2 \ell^i(\theta^*))$ and $s^i = (h^i)^{-1} \nabla f^i(\theta^*)$; then the asymptotic variance is $v^i = (h^i)^{-1} = \text{var}(s^i)$. Let $v^{12} = \text{cov}(s^1, s^2)$ be the correlation of the two estimators.

*Linear consensus with uniform weights*: $\hat{\theta}^{\text{linUnif}} = \frac{1}{2}(\hat{\theta}^1 + \hat{\theta}^2)$; the asymptotic variance is:

$$\text{var}\left(\frac{s^1 + s^2}{2}\right) = \frac{1}{4}(v^1 + v^2 + 2v^{12}).$$

*Linear consensus with Hessian weights* $w^i = h^i$: $\hat{\theta}^{\text{linHessian}} = (\hat{h}^1 + \hat{h}^2)^{-1}(\hat{h}^1 \hat{\theta}^1 + \hat{h}^2 \hat{\theta}^2)$. By Corollary 4.2, the asymptotic variance of $\hat{\theta}^{\text{linHessian}}$ is the same as that of $\hat{\theta}^{\text{joint}} = \arg\max_\theta \sum_i \ell^i(\theta)$, which is:

$$\text{var}(h^1 s^1 + h^2 s^2) = \frac{v^1 v^2 (v^1 + v^2 + 2v^{12})}{(v^1 + v^2)^2}.$$

*Linear consensus with optimal weights* $\hat{\theta}^{\text{linOpt}}$: By Proposition 4.6, the optimal weights for linear consensus are $w^{1*} = \frac{v^2 - v^{12}}{v^1 + v^2 - v^{12}}$ and $w^{2*} = \frac{v^2 - v^{12}}{v^1 + v^2 - v^{12}}$. The asymptotic variance is

$$\text{var}(w^{1*} s^1 + w^{2*} s^2) = \frac{v^1 v^2 - v^{12}}{v^1 + v^2 - 2v^{12}}.$$

*Max consensus with optimal weights* $\hat{\theta}^{\text{maxOpt}}$: By Proposition 4.4, for max consensus the weights $w^i = h^i$ are optimal. The asymptotic variance is $\min\{v^1, v^2\}$.

**Claim 4.9.** *In the toy case, we have $\hat{\theta}^{\text{linOpt}} \preceq \hat{\theta}^{\text{joint}} (= \hat{\theta}^{\text{linHessian}}) \preceq \hat{\theta}^{\text{linUnif}}$ and $\hat{\theta}^{\text{linOpt}} \preceq \hat{\theta}^{\text{maxOpt}}$, where $\hat{\theta}^a \preceq \hat{\theta}^b$ means $\text{MSE}(\hat{\theta}^a) \leq \text{MSE}(\hat{\theta}^b)$ asymptotically.*

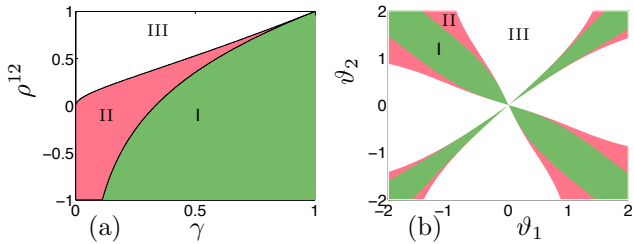

*Figure 1.* (a) Illustrating Claim 4.10: I (green): $\hat{\theta}^{\text{joint}} \preceq \hat{\theta}^{\text{linUnif}} \preceq \hat{\theta}^{\text{maxOpt}}$; II (red): $\hat{\theta}^{\text{joint}} \preceq \hat{\theta}^{\text{maxOpt}} \preceq \hat{\theta}^{\text{linUnif}}$; III (white): $\hat{\theta}^{\text{maxOpt}} \preceq \hat{\theta}^{\text{joint}} \preceq \hat{\theta}^{\text{linUnif}}$. (b) Comparing the algorithms when estimating $\theta$ (true $\theta^* = 1$) in a binary two-node model $p(x_1, x_2) \propto \exp(\theta x_1 x_2 + \vartheta_1 x_1 + \vartheta_2 x_2)$ as $\vartheta_1$ and $\vartheta_2$ (both known) are varied.

*Proof.* $\hat{\theta}^{\text{joint}} \preceq \hat{\theta}^{\text{linUnif}}$ is shown by the arithmetic-geometric mean inequality, and the rest since $\hat{\theta}^{\text{linHessian}}$ and $\hat{\theta}^{\text{maxOpt}}$ are special forms of linear consensus. □

However, $\hat{\theta}^{\text{linUnif}}$ or $\hat{\theta}^{\text{joint}}$ are not necessarily inferior or superior to $\hat{\theta}^{\text{maxOpt}}$. Their relative performance depends on the correlation and the quality (variance) of the two local estimators. Let $\rho^{12} = v^{12}/\sqrt{v^1 v^2}$ be the correlation coefficient of the estimators, and $\gamma = \min\{v^1/v^2, v^2/v^1\}$ the ratio of their variances.

**Claim 4.10.** *In the toy case, $\hat{\theta}^{\text{joint}}(= \hat{\theta}^{\text{linHessian}}) \preceq \hat{\theta}^{\text{maxOpt}}$ if and only if $\rho^{12} \leq \frac{1}{2}\sqrt{\gamma}(\gamma + 1)$. Similarly, $\hat{\theta}^{\text{linUnif}} \preceq \hat{\theta}^{\text{maxOpt}}$ if and only if $\rho^{12} \leq \frac{1}{2\sqrt{\gamma}}(3\gamma - 1)$;*

See Fig. 1 for illustration; this highlights the relative performance of max vs. linear consensus. While $\hat{\theta}^{\text{linHessian}}$ and $\hat{\theta}^{\text{linUnif}}$ tend to work better when the local estimators perform similarly ($\gamma \approx 1$) or when the local estimators have low or even negative correlations, $\hat{\theta}^{\text{maxOpt}}$ tends to work well when one local estimator is much better than the others ($\gamma \ll 1$) or when the local estimators are strongly positively correlated. This robustness makes max consensus useful for learning in difficult graphs, such as scale free graphs, for which standard methods often perform poorly (Ravikumar et al., 2010; Liu & Ihler, 2011). Fig. 1(b) illustrates how the values of $\gamma$ and $\rho^{12}$ are changed by varying the local potentials in a binary two-node model. Basically, $\hat{\theta}^{\text{maxOpt}}$ tends to work better when the magnitudes of the local potentials differ greatly, i.e., when the model is heteroskedastic.

## 5. Experiments

In this section, we test our algorithms on both small models (for which the asymptotic variance can be exactly calculated) and larger models of more practical size. We use a pairwise Ising model $p(x) \propto \exp(\sum_{(ij)\in E} \theta_{ij} x_i x_j + \sum_{i \in V} \theta_i x_i)$, $x_i \in \{-1, 1\}$, with random true parameters generated by $\theta_{ij} \sim \mathcal{N}(0, \sigma_{\text{pair}})$ and $\theta_i \sim \mathcal{N}(0, \sigma_{\text{singleton}})$. We test the Joint-MPLE, and the linear consensus with uniform weights (Linear-Uniform), with diagonal weights $\hat{w}_\alpha^i = 1/\hat{V}_{\alpha,\alpha}^i$ (Linear-Diagonal) and with the optimum vector weights given in Proposition 4.6 (Linear-Opt). We also test the max consensus with diagonal weights (Max-Diagonal(Opt)), which is optimal for max consensus (Proposition 4.4). We quantify the algorithms either by exactly calculated asymptotic efficiency, defined as $\text{tr}(V)/\text{tr}(V_{\text{mle}})$, or empirically by the MSE $||\hat{\theta} - \theta^*||^2$ calculated on simulated datasets.

### 5.1. Small Models

Two small graphs are considered: star graphs and grids, which have opposite topological properties. For these small models, we estimate the pairwise parameters $\theta_{ij}$ with known singleton parameters $\theta_i$.

**Star graphs.** A star graph has an unbalanced degree distribution, peaked at the center node. There has been theoretical and empirical work showing that such degree-unbounded graphs are harder to learn than more regular graphs (e.g., Liu & Ihler, 2011). From our perspective, the difficulty arises because the local estimators of high-degree nodes tend to deteriorate the overall performance. As we suggest in Section 4, the max consensus method is suitable for such graphs, as it can identify and discard the bad estimators.

The simulation supports this expectation. In Fig. 2(a), as degree increases, the variance of the local estimator of the center node increases quickly compared to the leaves (averaged). Fig. 2(b) shows the exact (solid lines) and empirical (dashed) asymptotic efficiencies of the algorithms on star graphs of different sizes. Linear-Uniform performs worst, since it fails to discount the influence of the worst estimator. Joint-MPLE and Linear-Diagonal perform better but still deteriorate as degree increases, since they downweight the worse estimators, but only to some extent. In contrast, Max-Diagonal is robust to the increasing degree, and can identify and discard the worst estimators. As theoretically predicted, Linear-Opt outperforms Max-Diagonal, but only slightly in this case. Note Linear-Opt is more costly than Max-Diagonal due to the extra communication step. The exact and empirical values in Fig. 2(b) match closely, showing the correctness of our theoretical analysis.

In Fig. 2(c) we show the effect of singleton potentials. The performance of one-step consensus methods generally decreases with higher magnitude singleton potentials, while the Joint-MPLE stays the same. Intuitively, this is because the local estimators are not able to jointly infer the local potentials, causing prob-lems when those local potentials dominate. Since our analysis is mainly asymptotic, we evaluate how the algorithms perform for small sample sizes in Fig. 2(d). As can be seen, the finite sample performance is essentially consistent with the asymptotic analysis.

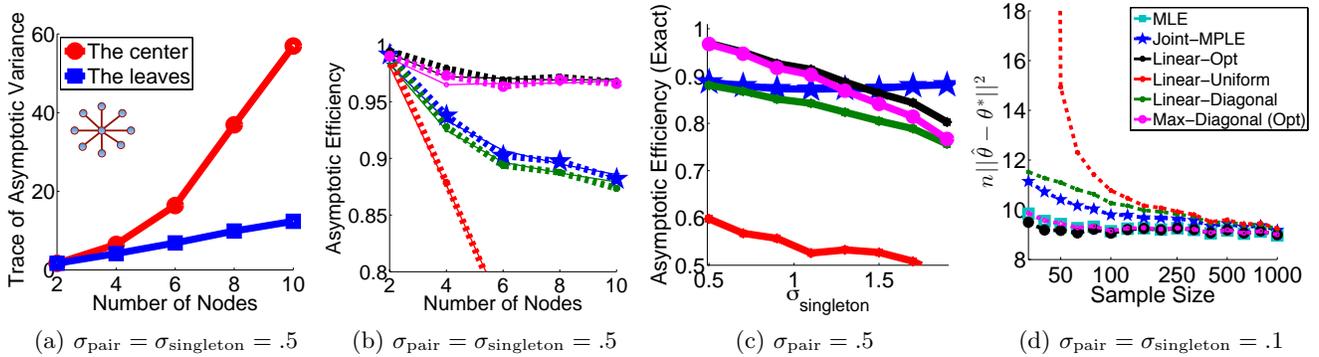

*Figure 2.* Results on star graphs. (a) Variance is much higher at the hub than at the leaves. (b) Exact (solid lines) and empirical (dashed lines) asymptotic efficiency of various algorithms vs. the size of the star graph. (c) The exact asymptotic efficiency for a 10-node star with $\sigma_{\text{pair}} = 0.5$ and $\sigma_{\text{singleton}} \in [.5, 2]$. (d). The mean square error vs. the number of data on a 10-node star graph; All the results are averaged on 50 random models, each with 50 random datasets.

**4×4 Grid.** The algorithms' performance on grids have the opposite trends; see Fig. 3(a). `Joint-MPLE` performs best, while `max-Diagonal` performs relatively poorly. This is because grids have balanced degree, and all the local estimators perform equally well. We check the finite sample performance of the algorithms in Fig. 3(b), which again shows similar trends to our asymptotic results. Finally, we show the convergence of ADMM in Fig. 3(c), illustrating that our initialization increases the convergence speed greatly.

### 5.2. Larger Models

We also test our algorithms on larger graphs, including a 100-node scale free network generated via the Barabási-Albert model (Barabási & Albert, 1999) and a 100-node Euclidean graph generated by connecting nearby sensors (distance ≤ .15) uniformly placed on the $[0, 1] \times [0, 1]$ plane; see Fig. 4. On these models, we estimate both the singleton and pairwise parameters. In Fig. 4(a)-(b) we see trends similar to their smaller analogues, the star graph and $4 \times 4$ grid, verifying that our analysis remains useful on models of larger sizes.

## 6. Related Work

A very recent, independently developed work (Wiesel & Hero, 2012) adopts a similar, but less general approach for Gaussian covariance estimation. They propose a similar linear consensus approach (using only uniform weights) and a similar parallel algorithm for joint MPLE, but do not discuss max consensus or linear consensus with general weights, and do not provide a comprehensive theoretical analysis. Another recent work (Eidsvik et al., 2010) uses composite likelihood for parallel computing on spatial data. Bradley & Guestrin (2011) gave a sample complexity analysis for MPLE and disjoint MPLE, which may be extensible to our algorithms.

Another line of work approximates MLE by estimating the partition function with variational algorithms (e.g., Wainwright, 2006; Sutton & McCallum, 2009). These methods can perform well at prediction tasks even with a "wrong" model, and can take a message-passing form potentially suitable to distributed settings. However, in terms of parameter estimation, these methods introduce a bias due to the approximate inference that is hard to estimate or control.

## 7. Conclusion

In this work, we present a general framework for distributed parameter learning. We show that the smart one-step consensus methods of the local estimators, especially those that exploit local second-order information, are both computationally efficient and statistically competitive with iterative methods using joint optimization. Particularly, we show that the max combination method is well suited to scale-free networks, a well-identified problem for existing methods. Our theory of combining estimators is quite general, and can be applied to other contexts to boost statistical performance. Future directions include considering model misspecification, finite sample complexity analysis and extension to high-dimensional structure learning.

**Acknowledgements.** Work supported in part by NSF IIS-1065618 and a Microsoft Research Fellowship.

<i>Distributed Parameter Estimation via Pseudo-likelihood</i>

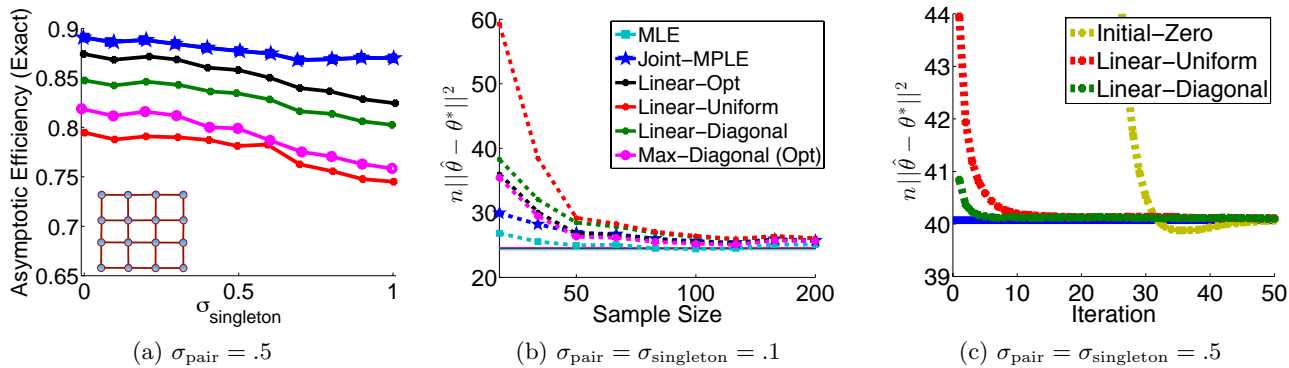

Figure 3. Results on $4 \times 4$ grid. (a) Exact asymptotic efficiency of the algorithms when $\sigma_{\text{singleton}} \in [0, 1]$. (b) Empirical MSE vs. data size. Solid horizontal lines show the theoretical asymptotic MSEs. (c) Convergence of ADMM, initialized at zero with $\rho_\alpha^i = 1$ (yellow), and initialized to linear consensus estimates with uniform (red) or diagonal (green) weights, with $\rho_\alpha^i$ set to the corresponding weights. All results are averaged on 50 random models with 50 datasets each.

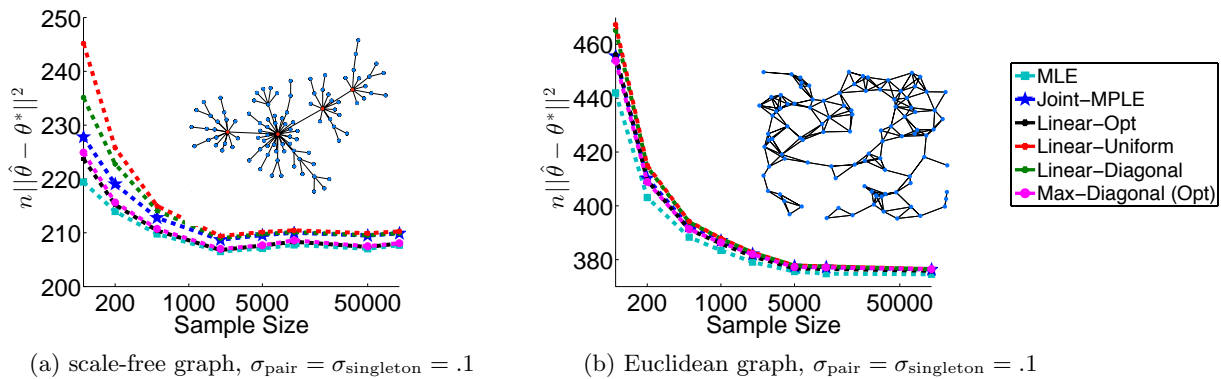

Figure 4. The empirical mean square error vs. the sample sizes on (a) a 100-node scale-free network and (b) a 100-node Euclidean graph. The results are averaged on 5 sets of random models and then 50 datasets, and show similar relative performance trends to the small-scale experiments.